\documentclass[runningheads]{llncs}

\usepackage{eccv}

\usepackage{eccvabbrv}

\usepackage[ruled,linesnumbered]{algorithm2e}
\usepackage{graphicx}
\usepackage{booktabs}

\usepackage[accsupp]{axessibility}  

\usepackage[pagebackref,breaklinks,colorlinks,citecolor=eccvblue]{hyperref}
 \usepackage{multirow}
\usepackage{orcidlink}

\usepackage{xcolor}
\definecolor{mycolorpurple}{RGB}{255, 102, 255}
\definecolor{mycolorred}{RGB}{227, 26, 26}
\definecolor{mycolorskyblue}{RGB}{6, 185, 238}
\usepackage{threeparttable}

\begin{document}

\title{GraphBEV: Towards Robust BEV Feature Alignment for Multi-Modal 3D Object Detection}

\titlerunning{Abbreviated paper title}

\author{Ziying Song\inst{1,2}\orcidlink{0000-0001-5539-2599} \and
Lei Yang\inst{3}\orcidlink{0000-0003-1800-6892} 
\and
Shaoqing Xu\inst{4}
\and
Lin Liu\inst{1,2}\orcidlink{0009-0005-3201-7610} 
\and
Dongyang Xu\inst{3}\orcidlink{0009-0008-9927-8649} 
\and
Caiyan Jia\inst{1,2}\orcidlink{0000-0003-0650-9564} \thanks{Corresponding author}
\and
Feiyang Jia\inst{1,2}\orcidlink{0000-0001-9933-872X}
\and
Li Wang\inst{5}\orcidlink{0000-0002-9325-2391}
}
\authorrunning{Z.~Song et al.}

\institute{School of Computer and Information Technology, Beijing Jiaotong University
\and
Beijing Key Lab of Traffic Data Analysis and Mining, China
\and 
School of Vehicle and Mobility, Tsinghua University
\and
Department of Electrome chanical Engineering, University of Macau
\and
School of Mechanical Engineering, Beijing Institute of Technology
\\
\email{\{songziying,cyjia\}@bjtu.edu.cn}\\
\url{https://github.com/adept-thu/GraphBEV}
}

\maketitle

\begin{abstract}
Integrating LiDAR and camera information into Bird's-Eye-View (BEV) representation has emerged as a crucial aspect of 3D object detection in autonomous driving. However, existing methods are susceptible to the inaccurate calibration relationship between LiDAR and the camera sensor. Such inaccuracies result in errors in depth estimation for the camera branch, ultimately causing misalignment between LiDAR and camera BEV features. In this work, we propose a robust fusion framework called GraphBEV. Addressing errors caused by inaccurate point cloud projection, we introduce a LocalAlign module that employs neighbor-aware depth features via Graph matching.
Additionally, we propose a GlobalAlign module to rectify the misalignment between LiDAR and camera BEV features. Our GraphBEV framework achieves state-of-the-art performance, with an mAP of 70.1\%, surpassing BEVFusion by 1.6\% on the nuScenes validation set. Importantly, our GraphBEV outperforms BEVFusion by 8.3\% under conditions with misalignment noise.  

\end{abstract}



\section{Introduction}
\label{sec:intro}
\begin{figure*}[t]
\centering
\includegraphics[width=1\textwidth]{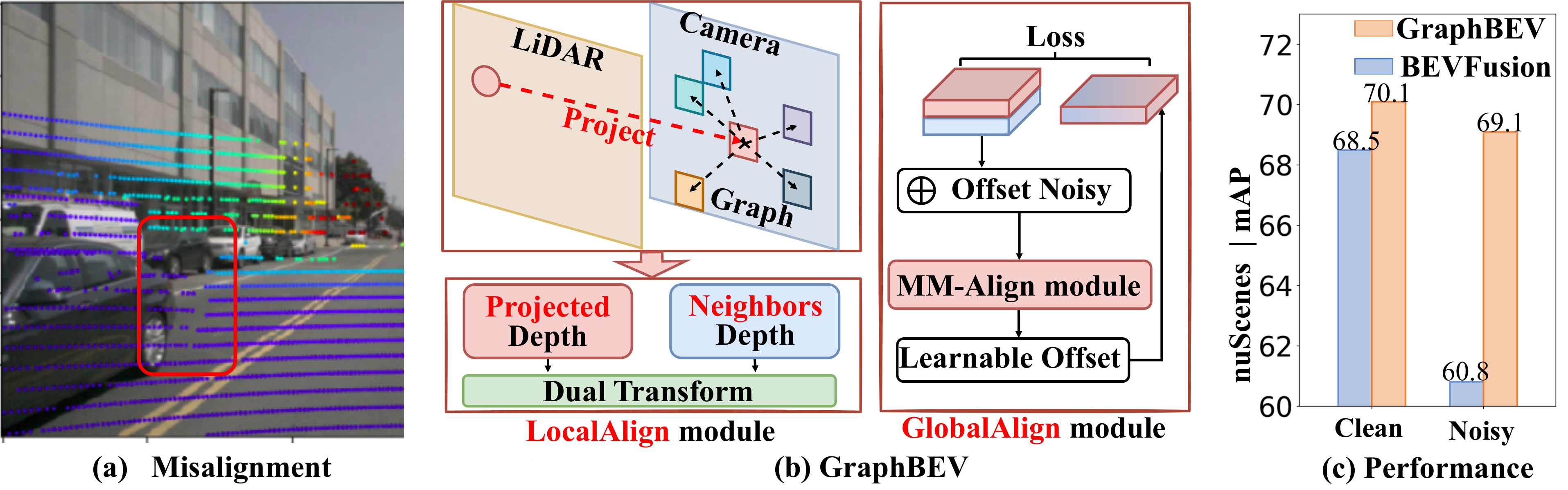}

\caption[ ]{\textbf{(a)} \textbf{Feature misalignment} primarily arises from overlooking the projection matrix errors between LiDAR and camera, leading to LiDAR-to-camera providing inaccurate depth for surrounding neighbors. \textbf{(b)} We propose GraphBEV, enhancing LiDAR-to-camera projected depth with neighboring depths through graph-based neighbor information construction for enriched contextual depth feature learning. Subsequently, we achieve the alignment of global multi-modal features by simulating LiDAR and camera BEV features' offsets and employing learnable offsets. \textbf{(c)} Empirical results reveal that our GraphBEV surpasses the BEVFusion \cite{bevfusion-mit} on the nuScenes by a margin of 1.6\% mAP on nuScenes validation dataset \cite{nuscenes} and by over 8.3\% on noisy misalignment settings \cite{zhujun_benchmarking}.}

\label{fig:motivation}
\end{figure*}

3D object detection is a critical component of autonomous driving systems, designed to accurately identify and localize objects such as cars, pedestrians, and other elements within a 3D environment \cite{song2024robustness,wang2023multi}. Current practices predominantly adhere to a multi-modal fusion paradigm such as BEVFusion \cite{bevfusion-mit,bevfusion-pku} for robust and high-quality detection. Different modalities often provide complementary information. For instance, images contain rich semantic representations but lack depth information. In contrast, point clouds offer geometric and depth information but are sparse and devoid of semantic information. Therefore, effectively leveraging the advantages of multi-modal data while mitigating their limitations is crucial for enhancing the robustness  of perception systems \cite{wang2023multi}.

Feature misalignment is a significant challenge in practical applications of multi-modal 3D object detection, primarily caused by calibration matrix errors between LiDAR and camera sensors \cite{yukaicheng_benchmarking,zhujun_benchmarking,song2024robustness}, as shown in Figure \ref{fig:motivation}(a). 
Multi-modal 3D object detection has evolved from early point-level \cite{mvx-net,pointpainting,epnet,epnet++,wang2021pointaugmenting,mvp} and feature-level \cite{VoxelNextFusion,autoalign,bi2024dyfusion,autoalignv2,graphalign,graphalign++}  methods to the currently prevalent methods of Bird's Eye View (BEV) fusion like BEVFusion \cite{bevfusion-pku,bevfusion-mit}. Although effective on clean datasets like nuScenes \cite{nuscenes}, the performance of BEVFusion\cite{bevfusion-mit} deteriorates on misaligned data, as illustrated in Figure \ref{fig:motivation}(c). This performance decline is primarily due to calibration errors between LiDAR and the camera, exacerbated by factors like road vibrations \cite{yukaicheng_benchmarking}. These inherent errors cannot be corrected through online calibration, presenting a significant challenge\cite{yukaicheng_benchmarking,zhujun_benchmarking}.

Most feature-level multi-modal methods \cite{DeepFusion,autoalign, Transfusion, DeepInteraction, CAT-Det} employ Cross Attention to query features of a specific modality, circumventing the need for projection matrices. A few feature-level multi-modal methods \cite{3dcvf,graphalign,graphalign++,autoalignv2, HMFI, Robust-FusionNet, Logonet,3DDualFusion} have sought to mitigate these errors through the use of projection offsets or neighboring projections. A few BEV-based methods, such as ObjectFusion \cite{ObjectFusion}, eliminate the camera-to-BEV transformation during fusion to align object-centric features across different modalities. MetaBEV \cite{ge2023metabev} utilizes Cross Deformable Attention for feature misalignment but overlooks depth estimation errors in view transformation, aligning features only during LiDAR and camera BEV fusion.

BEVFusion\cite{bevfusion-mit,bevfusion-pku} unites camera and LiDAR data in BEV space, enhancing detection but overlooks \textbf{feature misalignment} in real-world applications. It is primarily evident in two aspects: 1) BEVFusion\cite{bevfusion-mit} transforms multi-image features into a unified BEV representation using BEVDepth's\cite{bevdepth} explicit depth supervision from LiDAR-to-camera. While this LiDAR-to-camera strategy offers more reliable depth than LSS\cite{lss}, it overlooks the misalignment between LiDAR and camera in real-world scenarios, leading to \textbf{local misalignment}. 2) In the LiDAR-camera BEV fusion, misalignment of BEV features due to depth inaccuracies is overlooked by directly concatenating representations and applying basic convolution, as described in BEVFusion\cite{bevfusion-mit}, leading to \textbf{global misalignment}.

In this work, we propose a robust fusion framework, named GraphBEV to address the above feature misalignment problems. To tackle the local misalignment issue in camera-to-BEV transformation, we propose a LocalAlign module that first obtains neighbor depth information through a Graph in the View Transformation step of BEVFusion's \cite{bevfusion-mit} camera Branch, based on the explicit depth provided by the LiDAR-to-camera projection. Subsequently, we propose a GlobalAlign module that encodes the projected depth from LiDAR-to-camera and the neighboring depth through dual depth encoding to generate a new Reliable depth representation incorporating neighbor information. Furthermore, to address the global misalignment problem in the fusion of LiDAR and camera BEV features, we effectively resolve the global misalignment between the two BEV feature representations by dynamically generating offsets. To validate the effectiveness of GraphBEV, we evaluated it on the nuScenes dataset, a well-known benchmark for 3D object detection. Our GraphBEV not only achieves state-of-the-art (SOTA) performance under clean settings but also outperforms BEVFusion by over 8.3\% on non-aligned noisy settings nuScenes dataset provided by Dong et al. \cite{zhujun_benchmarking}. Notably, our GraphBEV experiences only a slight performance drop under noisy settings compared to clean settings. Notably, our GraphBEV bolsters the real-world applicability of BEV-based methods \cite{bevfusion-mit,bevfusion-pku} by addressing sensor calibration errors. Our contributions are summarized below:
\begin{enumerate}
\item 
We propose a robust fusion framework, named GraphBEV, to address feature misalignment arising from projection errors between LiDAR and camera inputs.
\item 
By deeply analyzing the fundamental causes of feature misalignment, we propose LocalAlign and GlobalAlign modules within our GraphBEV framework to address local misalignments from imprecise depth and global misalignments between LiDAR and camera BEV features.
\item 
Extensive experiments validate the effectiveness of our GraphBEV, demonstrating competitive performance on the nuScenes dataset. Notably, GraphBEV maintains comparable performance across both clean settings and misaligned noisy conditions.
\end{enumerate}
\section{Related Work}
\subsection{LiDAR-based 3D Object Detection}
LiDAR-based 3D object detection methods can be categorized into three primary types based on point cloud representation: Point-based, Voxel-based, and PV-based (Point-Voxel). 
Point-based methods \cite{lidarrcnn,Frustumpointnets,Pointnet,Pointnet++,Pointrcnn} extend PointNet's \cite{Pointnet,Pointnet++} principle, directly processing raw point clouds with stacked Multi-Layer Perceptrons (MLPs) to extract point features.
Voxel-based methods \cite{Voxelnet,Voxelrcnn,Second,LargeKernel3D,SATGCN,sparsedet} typically convert point clouds into voxels and apply 3D sparse convolutions for voxel feature extraction. In addition, PointPillars \cite{Pointpillars} converts irregular raw point clouds into pillars and encodes them on a 2D backbone, achieving a very high FPS. Some Voxel-based methods \cite{vsettransformer,voxeltransformer,Sparsetransformer,wang2024fuzzy} further exploit Transformers \cite{transformer} post-voxelization to capture long-range voxel relationships. PV-based methods \cite{vpnet,pvcnn,PVGNet,pvrcnn++,Std} combine voxel and point-based strategies, extracting features from point clouds' diverse representations using both approaches, achieving higher accuracy albeit with increased computational demand.

\subsection{Camera-based 3D Object Detection}
Camera-based 3D object detection methods have gained increasing attention in academia and industry, mainly due to the significantly lower cost of camera sensors compared to LiDAR \cite{song2024robustness}. Early methods \cite{brazil2019m3d,xu2018multi,simonelli2019disentangling} focused on augmenting 2D object detectors with additional 3D bounding box regression heads. Current camera-based methods have rapidly evolved since LSS \cite{lss} introduced the concept of unifying multi-view information onto a bird's-eye view (BEV) through `Lift, splat.' LSS-based methods \cite{lss,bevdepth,BEVStereo,park2022time,yang2023bevheight,yang2023bevheight++} like BEVDepth \cite{bevdepth} extract 2D features from multi-view images and provide effective depth supervision via LiDAR-to-camera projections before unifying multi-view features onto the BEV. Subsequent works \cite{BEVStereo,park2022time} have introduced multi-view stereo techniques to improve depth estimation accuracy and achieve SOTA performance. Additionally, inspired by the success of transformer-based architectures such as DETR \cite{detr} and Deformable DETR \cite{deformabledetr} in 2D detection, transformer-based detectors have emerged for 3D object detection. Following DETR3D \cite{wang2022detr3d}, these methods design a set of object queries \cite{jiang2023polarformer,liu2022petr,liu2023petrv2} or BEV grid queries \cite{bevformer,bevformerv2}, then perform view transformation through cross-attention between queries and image features.

\subsection{Multi-modal 3D Object Detection}
Multi-modal 3D object detection refers to using data features from different sensors and integrating these features to achieve complementarity, thus enabling the detection of 3D objects. Previous multi-model methods can be coarsely classified into three types by fusion, i.e., point-level, feature-level, and BEV-based methods. Point-level \cite{mvx-net,pointpainting,epnet,epnet++,wang2021pointaugmenting,mvp} and feature-level \cite{VoxelNextFusion,autoalign,bi2024dyfusion,autoalignv2,graphalign,graphalign++,zhang2023urformer,xu2024multifusionpatingtgrs,song2024robofusion} typically leverage image features to augment LiDAR points or 3D object proposals. BEV-based methods \cite{bevfusion-mit,bevfusion-pku,ge2023metabev, ObjectFusion,song2024contrastalign} efficiently unify the representations of LiDAR and camera into BEV space. Although BEVFusion\cite{bevfusion-mit,bevfusion-pku} achieves high performance, they are typically tested on clean datasets like nuScenes\cite{nuscenes}, overlooking real-world complexities, especially \textbf{feature misalignment}, which hampers their application.

\section{Method}
\begin{figure*}[t]
\centering
\includegraphics[width=1\textwidth]{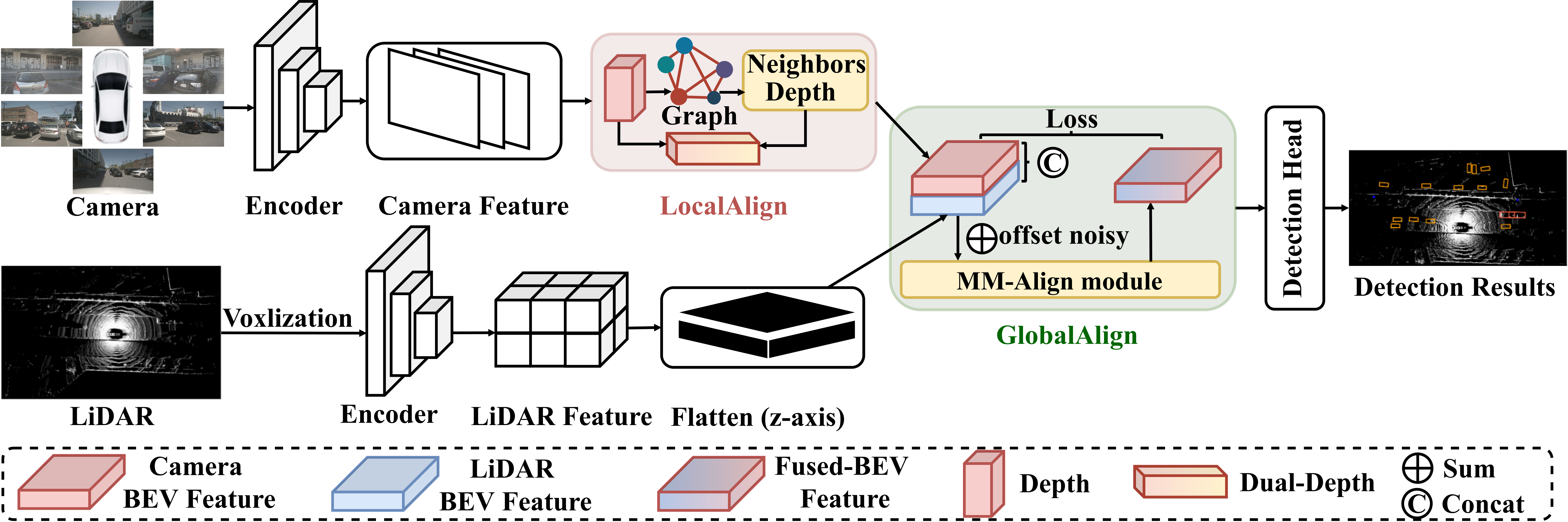}
\caption[ ]{The overview of \textbf{GraphBEV} framework. The LiDAR branch largely follows the baselines ~\cite{bevfusion-mit,Transfusion} to generate LiDAR BEV features. In the camera branch, we first extract camera BEV features using the proposed LocalAlign module, which aims to address local misalignment due to sensor calibration errors. Subsequently, we simulate the offset noise of LiDAR and camera BEV features, followed by aligning global multi-modal features through learnable offsets. Notably, we only add offset noise to the GlobalAlign module during \textbf{training} to simulate global misalignment issues. Finally, we employ a dense detection head \cite{Transfusion} to accomplish the 3D detection task.
}
\label{fig:framework}
\end{figure*}

To address the \textbf{feature misalignment} issue in previous BEV-based methods \cite{bevfusion-mit,bevfusion-pku}, we propose a robust fusion framework named \textbf{GraphBEV}. We provide an overview of our framework in Figure \ref{fig:framework}. Taking input from different sensors, including LiDAR and camera, we first apply modality-specific encoders, Swin-Transformer \cite{swimtransformer} as camera encoder and Second\cite{Second} as LiDAR encoder, to extract their respective features. Then, the camera features are transformed into the camera BEV feature through our proposed \textbf{LocalAlign} module, aiming to mitigate the local misalignment caused by the projection error between LiDAR and the camera in the camera-to-BEV process of previous BEV-based methods \cite{bevfusion-mit,bevfusion-pku}. The LiDAR features are then compressed along the Z-axis to represent 3D features as 2D LiDAR BEV features. Next, we propose a \textbf{GlobalAlign} module that further alleviates global misalignment between different modalities, including LiDAR and camera BEV features. Finally, we append detection head \cite{Transfusion} to accomplish the 3D detection task. Our baseline is the BEVFusion\cite{bevfusion-mit}, within which we detail the introduction of the LocalAlign and GlobalAlign modules below.

\subsection{LocalAlign}

\begin{figure}[t]
\centering
\includegraphics[width=1\textwidth]{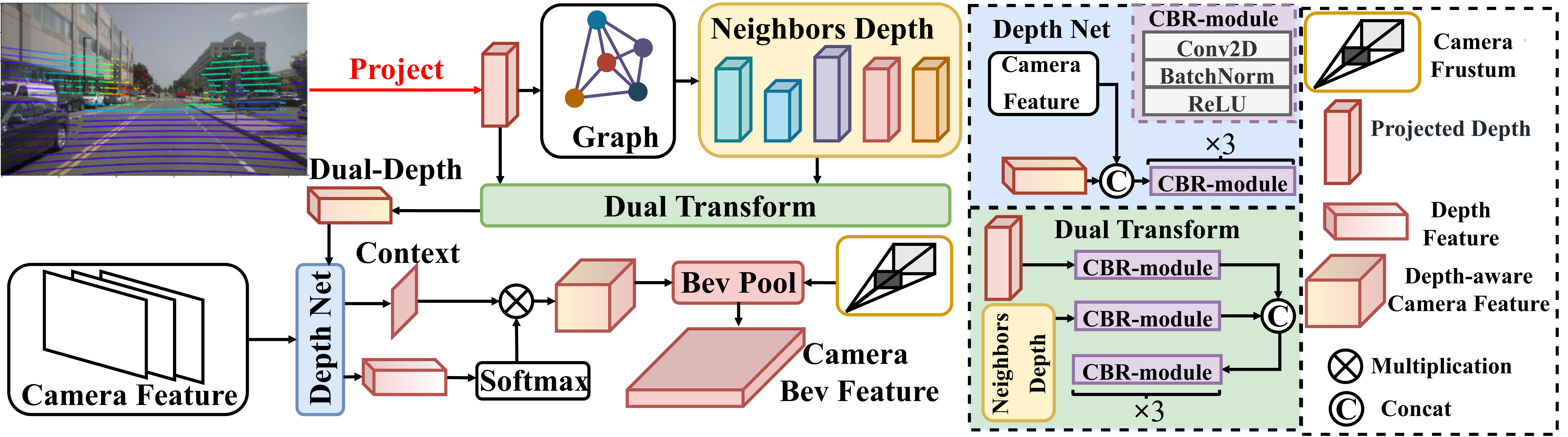}
\caption[ ]{The overview of \textbf{LocalAlign} pipeline. The LocalAlign module addresses Local Misalignment from LiDAR-to-camera by enhancing the camera-to-BEV Transform with neighboring depth features using a KD-Tree algorithm for nearest-neighbor relations.} 
\label{fig:localalign}
\end{figure}

To facilitate the transformation of camera features into BEV features, BEVFusion \cite{bevfusion-mit,bevfusion-pku} adopts the LSS-based\cite{lss} methods like BEVDepth \cite{bevdepth} leverages LiDAR-to-camera to provide projected depth, thereby enabling the fusion of depth and image features. In the process of camera-to-BEV, BEVFusion \cite{bevfusion-mit} and BEVDepth \cite{bevdepth}  operate under the assumption that the depth information provided by LiDAR-to-camera projection is accurate and reliable. However, they overlook the complexities inherent in real-world scenarios, where most of the projection matrices between LiDAR and cameras are calibrated manually. Such calibration inevitably introduces projection errors, leading to \textbf{depth misalignment}—where the depths of surrounding neighbors are projected as the pixel's depth. This depth misalignment results in inaccuracies within the depth features, causing misalignment during multi-view transformation into BEV representations. Given that LSS-based methods \cite{lss,bevdepth} rely on depth estimation from pixel-level features with inaccuracies in detailing, this leads to \textbf{local misalignment} within camera BEV features. It underscores the challenges of ensuring precise depth estimation within the BEVFusion \cite{bevfusion-mit} and highlights the importance of robust methods to address projection errors.

We propose a LocalAlign module to address Local Misalignment, with its pipeline depicted in Figure \ref{fig:localalign}. Specifically, LiDAR-to-camera provides projected depth, defined as \(D_{S} \in \mathbb{R}^{B_S \times N_{C} \times 1 \times H \times W}\), where \(B_S\) represents the batch size, \(N_{C}\) denotes the number of multi-views (six in the case of nuScenes), and \(H\) and \(W\) are the height and width of the images, respectively. The projection from LiDAR-to-camera maps 3D point clouds onto an image plane, from which we can obtain the indices of the projected pixels, defined as \(M_{\text{Coords}} \in \mathbb{R}^{N_{P} \times 2}\), where \(N_{P} \) refers to the number of points projected onto pixels, and 2 represents the pixel coordinates (\(u, v\)) as illustrated below.
\begin{equation}\label{equ3d2d}
z_{c}\left[\begin{array}{c}
u \\
v \\
1
\end{array}\right]=h \mathcal{K}\left[\begin{array}{ll}
R & T
\end{array}\right]\left[\begin{array}{c}
P_{x} \\
P_{y} \\
P_{z} \\
1
\end{array}\right]
\end{equation}
where, $P_{x}$, $P_{y}$, $P_{z}$ denote the LiDAR point's 3D locations, $u$, $v$ denote the corresponding 2D locations, and $z_{c}$ represents the depth of its projection on the image plane, $\mathcal{K}$ denotes the camera intrinsic parameter, $R$ and $T$ denote the rotation and the translation of the LiDAR with respect to the camera reference system, and $h$ denotes the scale factor due to down-sampling.

\begin{algorithm}[t]
\SetAlgoLined
\caption{Graph for Finding Neighbors} \label{algorithm:KD-Tree}
\KwIn{

The indices of the projected pixels \(M_{\text{Coords}} \in \mathbb{R}^{N_{P} \times 2}\).

Hyper-parameter: Number of neighbors $K_{\text{graph}} = 8$.
}
\KwOut{Neighbors $M_{K_{\text{Coords}}} \in \mathbb{R}^{N_{P} \times K_{\text{graph}} \times 2}$.
}

\While{LocalAlign}{
\SetKwProg{Fn}{Function}{}{}
\Fn{KD-Tree ($M_{\text{Coords}}$, ${M_{\text{Coords}}}_{i}$, $K_{\text{graph}}$)}{

        Compute the Euclidean distance between ${M_{\text{Coords}}}_{i}$ and $M_{\text{Coords}}$
        
        Indices = argsort(distances)
        
    \Return  $M_{\text{Coords}}[1:K_{\text{graph}}]$
}

\For{$i=1 \dots   N_{P} $}{
    Neighbors = KD-Tree($M_{\text{Coords}}$, ${M_{\text{Coords}}}_{i}$, $K_{\text{graph}}$)
    
    ${M_{\text{Coords}}}_{i}$ = Neighbors
}

}

\end{algorithm}

We employ the KD-Tree algorithm to obtain the indices of the projected pixels' neighbors, defined as $M_{K_{\text{Coords}}} \in \mathbb{R}^{N_{P} \times K_{\text{graph}} \times 2}$, where $K_{\text{graph}}$ denotes the number of neighbors for each projected pixel. The process is outlined in Algorithm (\ref{algorithm:KD-Tree}). It is worth noting that we simplified the process of the KD-Tree algorithm, and the code can be referred to in scipy\footnote{\url{https://github.com/minrk/scipy-1/blob/master/scipy/spatial/ckdtree.c}}. Then, we obtain the surrounding neighbor depth \(D_{K} \in \mathbb{R}^{B_S \times N_{C} \times K_{\text{graph}} \times H \times W} \) by indexing \(D_{S}\) with \(M_{\text{Coords}}\). Then, \(D_{S}\) and \(D_{K}\) simultaneously enter the Dual Transform module for deep feature encoding. The shapes of $D_{S}$ and $D_{K}$ are respectively modified to $[B_S \times N_{C} , 1, H, W]$ and $[B_S \times N_{C} , K_{\text{graph}} , H, W]$ before being fed into the Dual Transform module. This module comprises straightforward components, including convolutional layers, Batch Normalization, and ReLU activations, as illustrated in Figure \ref{fig:localalign}. The outcome of this process is the Dual Depth feature, denoted as $D_{\text{SK}}$, and its shape is $[B_S \times N_{C}, C_{\text{SK}}, \frac{H}{8}, \frac{W}{8}]$. The camera Encoder outputs multi-scale image features from the FPN, including $F_{\text{Cam.}} \in \mathbb{R}^{(B_S \times N_{C}) \times C_{\text{Cam}}\times \frac{H}{8} \times \frac{W}{8}}$ for richer semantic information and another at a reduced resolution of $\frac{H}{16}, \frac{W}{16}$. We opt to utilize the feature with the resolution $\frac{H}{8}, \frac{W}{8}$ due to its more comprehensive semantic content. 

The design of DepthNet is straightforward, as illustrated in Figure \ref{fig:localalign}. 
We input both $F_{\text{Cam.}}$ and $D_{\text{SK}}$ into DepthNet to fuse depth features with multi-view camera features. 
Initially, $F_{\text{Cam.}}$ and $D_{\text{SK}}$ are concatenated, followed by processing through three sets of CBR-module. 
This results in the generation of a new depth-aware camera feature, denoted as $F_{\text{DC}} \in \mathbb{R}^{(B_S \times N_{C}) \times C_{\text{DC}}\times \frac{H}{8} \times \frac{W}{8}}$. 
Subsequently, $F_{\text{DC}}$ is split along the $C_{\text{DC}}$ dimension into two new features: a novel depth feature, define as $\hat{F_{D}}  \in \mathbb{R}^{(B_S \times N_{C}) \times \hat{C_{D}}\times \frac{H}{8} \times \frac{W}{8}}$ and a novel image context feature, define as $\hat{F_{C}}  \in \mathbb{R}^{(B_S \times N_{C}) \times \hat{C_{C}} \times \frac{H}{8} \times \frac{W}{8}}$. 
It's important to note that $C_{\text{DC}}$ = $\hat{C_{C}} $ + $\hat{C_{D}}$, indicating the division of the combined feature space into distinct depth and image feature components. Subsequently, $\hat{F_{D}} $ is subjected to a softmax operation and then multiplied with $\hat{F_{C}} $, resulting in a novel image feature with depth information, represented as $\hat{F}_{\text{DC}} \in \mathbb{R}^{(B_S \times N_{C}) \times \hat{C_{C}}  \times \hat{C_{D}} \times \frac{H}{8} \times \frac{W}{8}}$. Finally, adopting operations consistent with LSS \cite{lss} and BEVDepth \cite{bevdepth}, we utilize pre-generated 3D spatial coordinates and \(\hat{F_{DC}}\) with BEV Pooling to output the camera BEV feature, thereby completing the camera-to-BEV transformation, and finally outputs the camera BEV feature, define as $ F^{C}_{B} \in \mathbb{R}^{B_S \times \hat{C_{C}}  \times H_{B} \times W_{B}}$.

\subsection{GlobalAlign}

\begin{figure}[t]
\centering
\includegraphics[width=0.75\textwidth]{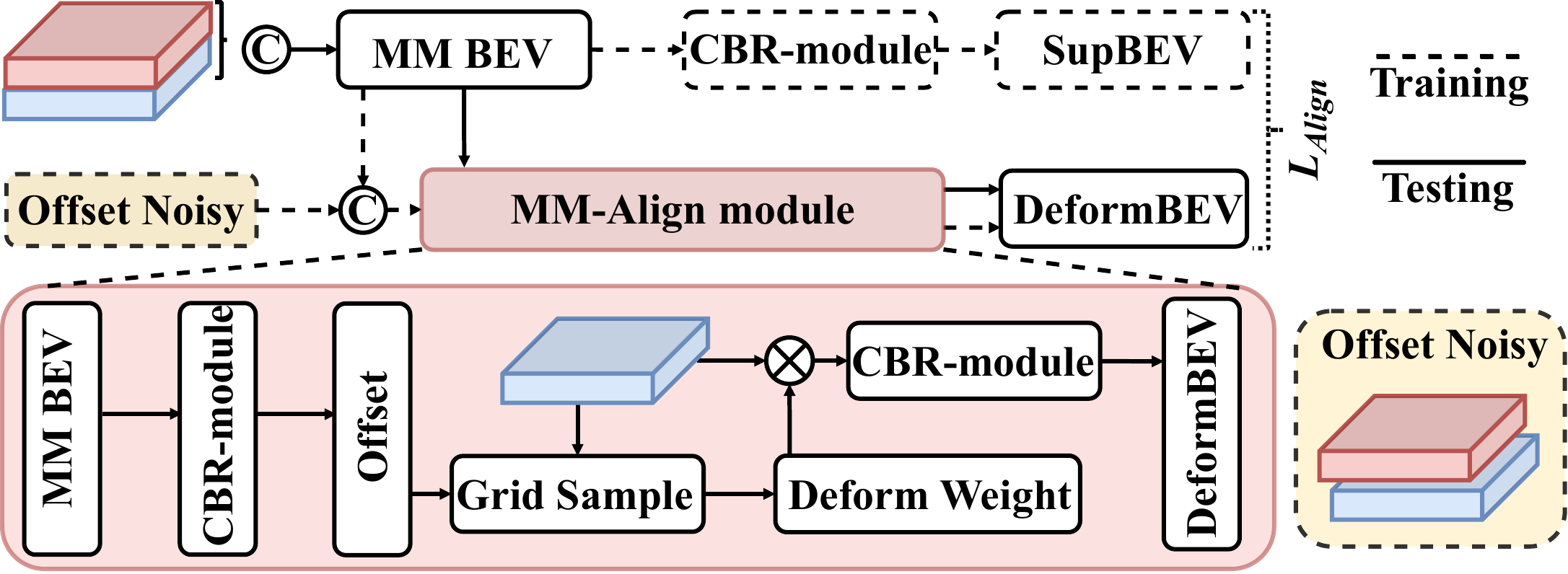}
\caption[ ]{The overview of \textbf{GlobalAlign} pipeline. The GlobalAlign module addresses the issue of misalignment in LiDAR-camera BEV feature fusion. During training, we add offset noise to simulate the global misalignment problem in the camera and LiDAR BEV features. It is supervised through a simple CBR-module to learn the offsets of camera BEV features. We do not introduce noise during testing and employ learnable offsets for forward inference.} 
\label{fig:globalalign}
\end{figure}

In the real world, feature misalignment due to calibration matrix discrepancies between LiDAR and camera sensors is inevitable. While the LocalAlign module mitigates local misalignment issues in the camera-to-BEV process, deviations may still exist in camera-BEV features. During LiDAR-camera BEV fusion, despite being in the same spatial domain, inaccuracies in depth from the view transformer and overlooking of global offsets between LiDAR BEV and camera BEV features lead to global misalignment.

To tackle the global misalignment issue described above, we introduce the GlobalAlign module, employing learnable offsets to achieve alignment of global multi-modal BEV features. As shown in Figures \ref{fig:framework} and \ref{fig:globalalign}, we use clean datasets such as nuScenes \cite{nuscenes} for training, which exhibit minimal deviations that can be considered negligible. The supervised information is derived from the features obtained after the fusion and convolution of LiDAR and camera BEV features. During training, we introduce global offset noise and employ learnable offsets. In the LiDAR branch, the LiDAR feature is flattened along the Z-axis to form the LiDAR BEV feature, defined as \( F^{L}_{B} \in \mathbb{R}^{B_S \times \hat{C_{L}}  \times H_{B} \times W_{B}} \). Initially, we concatenate \( F^{L}_{B} \) and \( F^{C}_{B} \) to obtain a fused BEV feature, denoted as \( F^{MM}_{B} \in \mathbb{R}^{B \times (\hat{C_{C}}  + \hat{C_{L}} ) \times H_{B} \times W_{B}} \). Subsequently, \( F^{MM}_{B} \) undergoes a convolution operation, resulting in a new fused feature, denoted as \( \hat{F_B} \in \mathbb{R}^{B_S \times \hat{C_{L}}  \times H_{B} \times W_{B}} \). Notably, \( \hat{F_B}\) will be utilized as a supervision signal during the training process.

As shown in Figure \ref{fig:globalalign}, we introduce random offset noise to the camera dimension of $ F^{MM}_{B}$ to obtain a new noisy feature $F^{MM}_{N} \in \mathbb{R}^{B_S \times (\hat{C_{C}}  + \hat{C_{L}} ) \times H_{B} \times W_{B}}$, simulating global misalignment issues originating from camera BEV features. Notably, the LiDAR BEV feature is directly flattened, thus more accurate. Then, $F^{MM}_{N}$ is input into the  MM-Align module for global offset learning. $F^{MM}_{N}$ is processed through the CBR-module with basic convolution operations to learn offsets, defined as $ F^{O} \in \mathbb{R}^{B_S \times 2 \times H_{B} \times W_{B}}$, where two refers to the offset coordinates $(u,v)$. Subsequently, LiDAR BEV features $ F^{L}_{B} $ and $ F^{O}$ undergo grid sampling to generate new deform weights, defined as $ F^{D}_{W} \in \mathbb{R}^{B_S \times \hat{C_{L}}  \times H_{B} \times W_{B}} $. The purpose of grid sampling is to utilize offsets for spatial transformation of LiDAR BEV feature $ F^{L}_{B}$, with learnable shifts dynamically adjusting to capture spatial dependencies more flexibly than standard convolution operations. Afterward, $ F^{D}_{W} $ is multiplied by LiDAR BEV features $ F^{L}_{B} $ to dynamically adjust features, followed by standard convolution operations through the CBR-module, culminating in the output Deform BEV defined as $ F^D_B \in \mathbb{R}^{B_S \times \hat{C_{L}} \times H_{B} \times W_{B}} $. Finally, during training, we supervise $F^D_B$ using $\hat{F_B}$ previously mentioned, and employ the $L_{\text{Align}}$ for supervision as follows:

\begin{equation}
L_{\text{Align}} = \frac{1}{N_{B}} \sum_{i=1}^{N_{B}} ({\hat{F_B}}_i - {F^D_B}_i)^2
\end{equation}
wherein, $N_{B} = B_S \times H_{B} \times W_{B}$ represents the total number of elements, and ${\hat{F_B}}_i$ and ${F^D_B}_i$ denote the value of the $i$th element in $\hat{F_B}$ and $F^D_B$, respectively. This formula calculates the mean of the squared differences between corresponding positions in the two feature maps, serving as the loss.

\section{Experiments}
In this section, we present the experimental setup of GraphBEV and evaluate the performance of 3D object detection on the nuScenes\cite{nuscenes} dataset. Additionally, we followed the solution provided by Ref.\cite{zhujun_benchmarking} to simulate feature misalignment scenarios.

\subsection{Experimental Setup}
\subsubsection{Dataset and Metric.}
We evaluate GraphBEV on the challenging large-scale nuScenes dataset \cite{nuscenes}, which was collected using a 32-beam LiDAR and six cameras. The dataset is typically split into 700/150/150 scenes for training, validation, and testing. The six images cover a 360-degree view of the surroundings, and the dataset provides calibration matrices that enable precise projection from 3D points to 2D pixels. We use mAP and NDS across all categories as the primary metrics for evaluation, following \cite{Transfusion,bevfusion-mit,bevfusion-pku}. Note that the NDS metric is a weighted average of mAP and other breakdown metrics (e.g., translation, scale, orientation, velocity, and attribute errors).
In addition, to validate the robustness of \textbf{feature alignment}, we followed Ref. \cite{zhujun_benchmarking} to simulate misalignment that LiDAR and camera projection metric errors. It is worth noting that Ref. \cite{zhujun_benchmarking} only added noise to the validation dataset, without the train and test datasets.

\subsubsection{Implementations.}
We implement GraphBEV within the PyTorch \cite{paszke2019pytorch}, built upon the open-source BEVFusion \cite{bevfusion-mit} and OpenPCDet \cite{openpcdet}. For the LiDAR branch, feature encoding is performed using SECOND \cite{Second} to obtain LiDAR BEV features, with voxel dimensions set to [0.075m, 0.075m, 0.2m] and point cloud ranges specified as [-54m, -54m, -5m, 54m, 54m, 3m] across the X, Y, and Z axes, respectively. The camera branch employs a Swin Transformer \cite{swimtransformer} as the backbone, integrating Heads of numbers 3, 6, 12, 24, and utilizing FPN \cite{maskrcnn} to fuse multi-scale feature maps. The resolution of input images is adjusted and cropped to 256 $\times$ 704. In the LSS \cite{lss} configuration, frustum ranges are set with X coordinates [-54m, 54m, 0.3m], Y coordinates: [-54m, 54m, 0.3m], Z coordinates [-10m, 10m, 20m], and depth range: [1m, 60m, 0.5m].

During training, we employ data augmentation for ten epochs, including random flips, rotations (within the range [$-\frac{\pi}{4}$, $\frac{\pi}{4}$]), translations (with std=0.5), and scaling in [0.9, 1.1] for LiDAR data enhancement. We use CBGS \cite{CBGS} to resample the training data. Additionally, we use random rotation in [$-5.4^\circ$, $5.4^\circ$] and random resizing in [0.38, 0.55] to augment the images. The Adam optimizer \cite{adam} is used with a one-cycle learning rate policy, setting the maximum learning rate to 0.001 and weight decay to 0.01. The batch size is 24, and training is conducted on 8 NVIDIA GeForce RTX 3090 24G GPUs. During inference, we remove Test Time Augmentation (TTA) data augmentation, and the batch size is set to 1 on an A100 GPU. All latency measurements are taken on the same workstation with an A100 GPU.

\subsection{Comparisons with State-of-the-Art Methods}

\begin{table*}[t]
\scriptsize
\centering
  \caption{Comparison with the SOTA methods on the nuScenes  \textcolor{blue}{validation} and \textcolor{red}{test}  set. `C.V.', `Motor.', `Ped.' and `T.C.' are short for construction vehicles, motorcycles, pedestrians, and traffic cones. The Modality column: `L' = only LiDAR data, `L.C.' = using both LiDAR and camera data. $^{\dagger}$ means using TTA (test-time augmentation). The best performances are marked with \textbf{bold} font.}
  \renewcommand\arraystretch{1}
  \resizebox{\linewidth}{!}{
  \begin{tabular}{l|c|cc|ccccc ccccc }
    \toprule
Method      &Modality       & mAP  & NDS  & Car  & Truck & C.V. & Bus  & Trailer & Barrier & Motor. & Bike & Ped. & T.C. \\ 
\midrule
\multicolumn{14}{c}{Performances on \textcolor{blue}{validation} set} \\
\midrule
TransFusion-L\cite{Transfusion} &L&  65.1 &70.1& 86.5& 59.6& 25.4& 74.4& 42.2& 74.1& 72.1& 56.0& 86.6& 74.1\\
FUTR3D\cite{chen2023futr3d}&LC& 64.2& 68.0& 86.3& 61.5& 26.0& 71.9& 42.1& 64.4& 73.6& 63.3& 82.6& 70.1\\
TransFusion\cite{Transfusion}& LC & 67.3 &71.2& 87.6& 62.0& 27.4& 75.7& 42.8& 73.9& 75.4& 63.1& 87.8& 77.0\\ 
BEVFusion-PKU \cite{bevfusion-pku} &LC & 67.9& 71.0& 88.6& 65.0& 28.1& 75.4& 41.4& 72.2& 76.7& 65.8& 88.7& 76.9\\
ObjectFusion \cite{ObjectFusion}& LC &69.8& 72.3 &89.7& \textbf{65.6}& \textbf{32.0}& \textbf{77.7}& 42.8& 75.2& 79.4& 65.0& 89.3& 81.1\\
\midrule
BEVFusion-MIT \cite{bevfusion-mit}& LC& 68.5 & 71.4 & 89.2 & 64.6 & 30.4 & 75.4 & 42.5 & 72.0 & 78.5 & 65.3 & 88.2 & 79.5\\ 
\textbf{GraphBEV(Ours)}& LC   &\textbf{70.1}&\textbf{72.9}& \textbf{89.9}& 64.7& 31.1& 76.0& \textbf{43.8}& \textbf{76.0}& \textbf{80.1}& \textbf{67.5}& \textbf{89.2} &\textbf{82.2}\\ 
 &&\textit{\fontsize{6}{0}\selectfont\textcolor{blue}{+1.6}}&\textit{\fontsize{6}{0}\selectfont\textcolor{blue}{+1.5}} &&& &&&\textit{\fontsize{6}{0}\selectfont\textcolor{blue}{+4.0}} &&\textit{\fontsize{6}{0}\selectfont\textcolor{blue}{+2.2}}&\textit{\fontsize{6}{0}\selectfont\textcolor{blue}{+2.7}}\\
\midrule
\midrule
\multicolumn{14}{c}{Performances on \textcolor{red}{test} set} \\
\midrule
PointPillar \cite{Pointpillars} & L & 40.1 & 55.0 & 76.0 & 31.0 & 11.3 & 32.1 & 36.6 & 56.4 & 34.2 & 14.0 & 64.0 & 45.6\\
CenterPoint \cite{centerpoint}$^{\dagger}$ &L& 60.3 & 67.3 & 85.2 & 53.5& 20.0 & 63.6 & 56.0 & 71.1& 59.5& 30.7 & 84.6& 78.4\\
PointPainting \cite{pointpainting}& LC       & 46.4 & 58.1 & 77.9 & 35.8  & 15.8 & 36.2 & 37.3    & 60.2    & 41.5   & 24.1 & 73.3 & 62.4 \\
PointAugmenting\cite{wang2021pointaugmenting}$^{\dagger}$ & LC &66.8& 71.0& 87.5& 57.3& 28.0& 65.2 &60.7& 72.6& 74.3& 50.9& 87.9& 83.6\\
MVP  \cite{mvp}& LC                 & 66.4 & 70.5 & 86.8 & 58.5  & 26.1 & 67.4 & 57.3    & 74.8    & 70.0   & 49.3 & 89.1 & 85.0 \\
GraphAlign \cite{graphalign}& LC  &66.5&70.6& 87.6& 57.7& 26.1& 66.2& 57.8& 74.1& 72.5& 49.0& 87.2 &86.3\\
AutoAlignV2\cite{autoalignv2}& LC & 68.4 & 72.4 & 87.0 & 59.0 & 33.1 & 69.3 & 59.3 & - & 72.9 & 52.1 & 87.6 & -\\
TransFusion \cite{Transfusion}& LC & 68.9 & 71.7 & 87.1 & 60.0 & 33.1 & 68.3 & 60.8 & 78.1 & 73.6 & 52.9 & 88.4 & 86.7\\
DeepInteraction\cite{DeepInteraction} &LC  &70.8& 73.4& 87.9& 60.2& 37.5& 70.8& 63.8& 80.4& 75.4& \textbf{54.5}& 90.3& 87.0 \\
BEVFusion-PKU  \cite{bevfusion-pku}& LC & 69.2 & 71.8 & 88.1 & \textbf{60.9} & 34.4 & 69.3 & 62.1 & 78.2 & 72.2 & 52.2 & 89.2 & 85.2\\
ObjectFusion \cite{ObjectFusion}& LC &71.0& 73.3& \textbf{89.4}& 59.0 & 40.5 & 71.8 & 63.1& 76.6 & \textbf{78.1} & 53.2 & 90.7 & 87.7\\

\midrule
BEVFusion-MIT \cite{bevfusion-mit}& LC & 70.2 & 72.9 & 88.6 & 60.1 & 39.3& 69.8 & 63.8 & 80.0 & 74.1 & 51.0 & 89.2 & 86.5\\ 
\textbf{GraphBEV(Ours)}& LC            & \textbf{71.7} & \textbf{73.6}& 89.2& 60.0 & \textbf{40.8} & \textbf{72.1} & \textbf{64.5}& \textbf{80.1} & 76.8 & 53.3 & \textbf{90.9}& \textbf{88.9}\\
&&\textit{\fontsize{6}{0}\selectfont\textcolor{red}{+1.5}}&\textit{\fontsize{6}{0}\selectfont\textcolor{red}{+0.7}} &&& &\textit{\fontsize{6}{0}\selectfont\textcolor{red}{+2.3}}&& &\textit{\fontsize{6}{0}\selectfont\textcolor{red}{+2.7}}&\textit{\fontsize{6}{0}\selectfont\textcolor{red}{+2.3}}&&\textit{\fontsize{6}{0}\selectfont\textcolor{red}{+2.4}}
\\
\bottomrule
\end{tabular} }
\label{tab_nuscenes_val_test}
\end{table*}

\begin{table*}[htp]
\scriptsize
\centering
  \caption{Comparison with the SOTA methods on BEV map segmentation on nuScenes \textcolor{blue}{validation} set. The Modality column: `L' = only LiDAR data, `LC' = using both LiDAR and camera data.}
  \renewcommand\arraystretch{1}
  \resizebox{\linewidth}{!}{
  \begin{tabular}{l|c|ccc cccc }
    \toprule
Method      &Modality & Drivable & Ped. Cross. & Walkway & Stop Line & Carpark & Divider & Mean \\ 
\midrule

PointPillars \cite{Pointpillars} & L & 72.0 & 43.1 & 53.1 & 29.7 & 27.7 & 37.5 & 43.8\\
CenterPoint \cite{centerpoint} & L & 75.6 &  48.4 & 57.5 & 36.5 & 31.7 & 41.9 & 48.6\\
PointPainting \cite{pointpainting} & LC & 75.9 & 48.5 & 57.1 & 36.9 & 34.5 & 41.9 & 49.1\\
MVP\cite{mvp} &  LC & 76.1 & 48.7 & 57.0 & 36.9 & 33.0 & 42.2 & 49.0 \\
\midrule
BEVFusion\cite{bevfusion-mit}& LC & 85.5 & 60.5 & 67.6 & 52.0 & 57.0 & \textbf{53.7} & 62.7\\ 
\textbf{GraphBEV(Ours)}& LC & \textbf{86.3} & \textbf{60.9} & \textbf{69.1} & \textbf{53.1} & \textbf{57.5} & 53.1 & \textbf{63.3}\\ 
\bottomrule
\end{tabular} }

\label{tab_nuscenes_bevmap}
\end{table*}

\subsubsection{3D Object Detection.}
We first compare the performances of our GraphBEV and other SOTA methods on the nuScenes \textcolor{blue}{validation} and \textcolor{red}{test} for 3D object detection task, as shown in Table \ref{tab_nuscenes_val_test}. Our GraphBEV achieves the best performances on the validation set (70.1\% in mAP and 72.9\% in NDS), which consistently outperforms all single-modality and multi-modal fusion methods.
Among the multi-model methods, PointPainting\cite{pointpainting}, PointAugmenting\cite{wang2021pointaugmenting}, and MVP\cite{mvp}, serving as point-level methods, are unable to avoid feature misalignment issues. On the other hand, GraphAlign\cite{graphalign}, AutoAlignV2\cite{autoalignv2}, TransFusion\cite{Transfusion}, and DeepInteraction \cite{DeepInteraction}, as feature-level methods, address the misalignment between LiDAR and camera features. However, these methods do not include the camera-to-BEV process, thus they cannot resolve the BEVFusion's \cite{bevfusion-mit,bevfusion-pku} feature misalignment issue caused by depth estimation.

As shown in Table \ref{tab_nuscenes_val_test}, our GraphBEV outperforms the baseline BEVFusion\cite{bevfusion-mit} by +1.6\% in mAP and +1.5\% in NDS. The open-source nuScenes dataset generally has minimal feature alignment issues but cannot entirely avoid them, as illustrated in Figure \ref{fig:motivation}(a). Specifically, our GraphBEV utilizes the KD-Tree algorithm to search the neighbor depth around the projected depth provided by LiDAR-to-camera. Compared to BEVFusion\cite{bevfusion-mit}, our GraphBEV demonstrates significant improvements in small objects, such as Barrier by +4.0\%, Bike by +2.2\%, and Pedestrian by +2.7\%. It is primarily caused by small objects being more sensitive to feature misalignment, where even slight misalignment between LiDAR and the camera can result in greater misalignment in small objects. Furthermore, our GraphBEV and ObjectFusion tackle the issue of feature misalignment in BEV-based multi-modal methods \cite{bevfusion-mit,bevfusion-pku}. While ObjectFusion\cite{ObjectFusion} introduces a post-fusion paradigm utilizing RoI Pooling fusion, our GraphBEV does not alter the paradigm of BEVFusion\cite{bevfusion-mit,bevfusion-pku}, slightly outperforms ObjectFusion by 0.7\% in mAP and 0.3\% in NDS.
Additionally,  our GraphBEV achieved SOTA performance when we submitted the detection results for the nuScenes test set to the official evaluation server. It surpassed the baseline BEVFusion\cite{bevfusion-mit} by 1.5\%  in mAP and 0.7\%  in NDS. Overall, we could address feature misalignment issues in BEVFusion\cite{bevfusion-mit,bevfusion-pku} without altering its excellent paradigm.

\subsubsection{BEV map segmentation.}
To evaluate 3D object detection, we also assessed the generalization capability in the BEV Map Segmentation (semantic segmentation) task on the nuScenes validation set, as shown in Table \ref{tab_nuscenes_bevmap}. Following the same training strategy as the baseline BEVFusion, we conducted evaluations within the [-50m, 50m]×[-50m, 50m] region around the ego car for each frame. We reported Intersection-over-Union (IoU) scores for drivable area, pedestrian crossing, walkway, stop line, car park, and divider. Significant improvements were observed for drivable area, pedestrian crossing, walkway, stop line, and car park, with only a minor decrease observed for the divider. Overall, our GraphBEV demonstrates not only significant performance in 3D object detection but also strong generalization capability in BEV Map Segmentation.

\subsection{Ablation Study}

\begin{table*}[t]
\scriptsize
\centering
  \caption{Roles of Different Modules in GraphBEV for Feature Alignment on nuScenes validation set under \textcolor{blue}{clean} setting and \textcolor{red}{noisy} misalignment setting. `C.V.', `Motor.', `Ped.' and `T.C.' are short for construction vehicles, motorcycles, pedestrians, and traffic cones.
 `+L (only)' indicates the addition of only the LocalAlign module, and `+G (only)' indicates only the GlobalAlign module. GraphBEV denotes the addition of both LocalAlign and GlobalAlign modules. `L.T. (ms)' represents latency. All latency measurements are conducted on the same workstation with an A100 GPU.
  }
  \renewcommand\arraystretch{1}
  \resizebox{\linewidth}{!}{
  \begin{tabular}{l|c|ccc|ccccc ccccc }
    \toprule
 &Method            & mAP  & NDS & LT(ms)  & Car  & Truck & C.V. & Bus  & Trailer & Barrier & Motor. & Bike & Ped. & T.C. \\ 
\cmidrule(r){2-15}
&TransFusion\cite{Transfusion}& 67.3& 71.2&164.6 &87.6& 62.0& 27.4& 75.7& 42.8& 73.9& 75.4& 63.1& 87.8& 77.0\\
\cmidrule(r){2-15}
\multirow{7}{*}{\textcolor{blue}{\rotatebox{90}{\textbf{Clean}}}} &Baseline \cite{bevfusion-mit}&  68.5 & 71.4 &133.2 & 89.2 & 64.6 & 30.4 & 75.4 & 42.5 & 72.0 & 78.5 & 65.3 & 88.2 & 79.5\\
\cmidrule(r){2-15}
&\multirow{2}{*}{+L (only)}& 69.7&72.4&136.3& 89.5& 64.4& 30.6& 75.9& 43.5& 75.6& 79.6& 67.1& 88.8 &82.3\\
 &
 &\textit{\fontsize{6}{0}\selectfont\textcolor{blue}{+1.2}}
 &\textit{\fontsize{6}{0}\selectfont\textcolor{blue}{+1.0}}
 &\textit{\fontsize{6}{0}\selectfont\textcolor{blue}{+3.1}}
 &\textit{\fontsize{6}{0}\selectfont\textcolor{blue}{+0.3}}
 &\textit{\fontsize{6}{0}\selectfont\textcolor{blue}{-0.2}}
 &\textit{\fontsize{6}{0}\selectfont\textcolor{blue}{+0.2}}
 &\textit{\fontsize{6}{0}\selectfont\textcolor{blue}{+0.5}}
 &\textit{\fontsize{6}{0}\selectfont\textcolor{blue}{+1.0}}
 &\textit{\fontsize{6}{0}\selectfont\textcolor{blue}{+3.6}} 
 &\textit{\fontsize{6}{0}\selectfont\textcolor{blue}{+1.1}} 
 &\textit{\fontsize{6}{0}\selectfont\textcolor{blue}{+1.8}}
 &\textit{\fontsize{6}{0}\selectfont\textcolor{blue}{+0.6}}
 &\textit{\fontsize{6}{0}\selectfont\textcolor{blue}{+2.8}}
 \\
\cmidrule(r){2-15}
&\multirow{2}{*}{+G (only)}&  68.9 & 71.7 &138.1& 89.6 & 64.7 & 30.5 & 75.7 & 43.4 & 72.2 & 79.2 & 65.8 & 88.7 & 79.9\\
 &
 &\textit{\fontsize{6}{0}\selectfont\textcolor{blue}{+0.4}}
 &\textit{\fontsize{6}{0}\selectfont\textcolor{blue}{+0.3}}
 &\textit{\fontsize{6}{0}\selectfont\textcolor{blue}{+4.9}}
 &\textit{\fontsize{6}{0}\selectfont\textcolor{blue}{+0.4}}
 &\textit{\fontsize{6}{0}\selectfont\textcolor{blue}{+0.1}}
 &\textit{\fontsize{6}{0}\selectfont\textcolor{blue}{+0.1}}
 &\textit{\fontsize{6}{0}\selectfont\textcolor{blue}{+0.3}}
 &\textit{\fontsize{6}{0}\selectfont\textcolor{blue}{+0.9}}
 &\textit{\fontsize{6}{0}\selectfont\textcolor{blue}{+0.2}} 
 &\textit{\fontsize{6}{0}\selectfont\textcolor{blue}{+0.7}} 
 &\textit{\fontsize{6}{0}\selectfont\textcolor{blue}{+0.5}}
 &\textit{\fontsize{6}{0}\selectfont\textcolor{blue}{+0.5}}
 &\textit{\fontsize{6}{0}\selectfont\textcolor{blue}{+0.4}}
 \\
\cmidrule(r){2-15}
&\textbf{GraphBEV}&   70.1&72.9&140.9& 89.9& 64.7& 31.1& 76.0& 43.8& 76.0& 80.1& 67.5& 89.2 &82.2\\ 
 &
 &\textit{\fontsize{6}{0}\selectfont\textcolor{blue}{+1.6}}
 &\textit{\fontsize{6}{0}\selectfont\textcolor{blue}{+1.5}}
 &\textit{\fontsize{6}{0}\selectfont\textcolor{blue}{+7.7}}
 &\textit{\fontsize{6}{0}\selectfont\textcolor{blue}{+0.7}}
 &\textit{\fontsize{6}{0}\selectfont\textcolor{blue}{+0.1}}
 &\textit{\fontsize{6}{0}\selectfont\textcolor{blue}{+0.7}}
 &\textit{\fontsize{6}{0}\selectfont\textcolor{blue}{+0.6}}
 &\textit{\fontsize{6}{0}\selectfont\textcolor{blue}{+1.3}}
 &\textit{\fontsize{6}{0}\selectfont\textcolor{blue}{+4.0}} 
 &\textit{\fontsize{6}{0}\selectfont\textcolor{blue}{+1.6}}
 &\textit{\fontsize{6}{0}\selectfont\textcolor{blue}{+2.2}}
 &\textit{\fontsize{6}{0}\selectfont\textcolor{blue}{+1.0}}
 &\textit{\fontsize{6}{0}\selectfont\textcolor{blue}{+2.7}}
 \\
\midrule
\midrule
&TransFusion\cite{Transfusion}& 66.4& 70.6&164.6 &86.3& 61.8& 26.9& 75.1& 42.0& 73.1& 74.9& 62.5& 85.2& 75.9\\
\cmidrule(r){2-15}
\multirow{8}{*}{\textcolor{red}{\rotatebox{90}{\textbf{Noisy}}}}&Baseline\cite{bevfusion-mit}&  60.8 & 65.7 &132.9 & 83.1 & 50.3 & 26.5& 66.4 & 38.0 & 65.0 & 64.9 & 52.8 & 86.1 & 75.1\\ 
\cmidrule(r){2-15}
&\multirow{2}{*}{+L (only)}&   67.0&70.1&136.2& 86.4& 60.3& 29.1& 73.3& 40.3& 74.0& 78.0& 62.1& 86.8 &79.9\\ 
 &&\textit{\fontsize{6}{0}\selectfont\textcolor{red}{+6.2}}
 &\textit{\fontsize{6}{0}\selectfont\textcolor{red}{+4.4}}
 &\textit{\fontsize{6}{0}\selectfont\textcolor{red}{+3.3}}
 &\textit{\fontsize{6}{0}\selectfont\textcolor{red}{+3.3}} 
 &\textit{\fontsize{6}{0}\selectfont\textcolor{red}{+10.0}} 
 &\textit{\fontsize{6}{0}\selectfont\textcolor{red}{+2.6}}
 &\textit{\fontsize{6}{0}\selectfont\textcolor{red}{+6.9}}
 &\textit{\fontsize{6}{0}\selectfont\textcolor{red}{+2.3}}
 &\textit{\fontsize{6}{0}\selectfont\textcolor{red}{+9.0}} 
 &\textit{\fontsize{6}{0}\selectfont\textcolor{red}{+13.1}} 
 &\textit{\fontsize{6}{0}\selectfont\textcolor{red}{+9.3}}
 &\textit{\fontsize{6}{0}\selectfont\textcolor{red}{+0.7}}
 &\textit{\fontsize{6}{0}\selectfont\textcolor{red}{+4.8}}
 \\
 \cmidrule(r){2-15}
 &\multirow{2}{*}{+G (only)}&    63.1&67.2&137.9& 84.2& 51.7& 27.8& 68.6& 39.5& 68.8& 68.7& 57.2& 86.2& 77.8\\  
 &&\textit{\fontsize{6}{0}\selectfont\textcolor{red}{+2.3}}
 &\textit{\fontsize{6}{0}\selectfont\textcolor{red}{+1.5}}
 &\textit{\fontsize{6}{0}\selectfont\textcolor{red}{+5.0}}
 &\textit{\fontsize{6}{0}\selectfont\textcolor{red}{+1.1}} 
 &\textit{\fontsize{6}{0}\selectfont\textcolor{red}{+1.4}} 
 &\textit{\fontsize{6}{0}\selectfont\textcolor{red}{+1.3}}
 &\textit{\fontsize{6}{0}\selectfont\textcolor{red}{+2.2}}
 &\textit{\fontsize{6}{0}\selectfont\textcolor{red}{+1.5}}
 &\textit{\fontsize{6}{0}\selectfont\textcolor{red}{+3.8}} 
 &\textit{\fontsize{6}{0}\selectfont\textcolor{red}{+3.8}} 
 &\textit{\fontsize{6}{0}\selectfont\textcolor{red}{+4.4}}
 &\textit{\fontsize{6}{0}\selectfont\textcolor{red}{+0.1}}
 &\textit{\fontsize{6}{0}\selectfont\textcolor{red}{+2.7}}
 \\
 \cmidrule(r){2-15}
 &\textbf{GraphBEV}&   69.1&72.0&141.0& 88.1& 63.5& 30.0& 75.1& 42.7& 75.3& 79.8& 64.9& 88.9 &82.2\\ 
 &&\textit{\fontsize{6}{0}\selectfont\textcolor{red}{+8.3}}
 &\textit{\fontsize{6}{0}\selectfont\textcolor{red}{+6.3}}
 &\textit{\fontsize{6}{0}\selectfont\textcolor{red}{+8.1}}
 &\textit{\fontsize{6}{0}\selectfont\textcolor{red}{+5.0}} 
 &\textit{\fontsize{6}{0}\selectfont\textcolor{red}{+13.2}} 
 &\textit{\fontsize{6}{0}\selectfont\textcolor{red}{+3.5}}
 &\textit{\fontsize{6}{0}\selectfont\textcolor{red}{+8.7}}
 &\textit{\fontsize{6}{0}\selectfont\textcolor{red}{+4.7}}
 &\textit{\fontsize{6}{0}\selectfont\textcolor{red}{+10.3}} 
 &\textit{\fontsize{6}{0}\selectfont\textcolor{red}{+14.9}} 
 &\textit{\fontsize{6}{0}\selectfont\textcolor{red}{+12.1}}
 &\textit{\fontsize{6}{0}\selectfont\textcolor{red}{+2.8}}
 &\textit{\fontsize{6}{0}\selectfont\textcolor{red}{+7.1}}
 \\
\bottomrule
\end{tabular} }

\label{tab_nuscenes_misalignment}
\end{table*}




\subsubsection{Roles of Different Modules in GraphBEV for Feature Alignment.}
To analyze the impact of misalignment, we conducted comparative experiments between our GraphBEV and BEVFusion \cite{bevfusion-mit}. It is noteworthy that in Table \ref{tab_nuscenes_misalignment}, we introduced misalignment into the nuScenes validation set, rather than in the training and testing sets, as follow  Ref. \cite{zhujun_benchmarking}. We trained on the clean nuScenes \cite{nuscenes} training dataset and evaluated performance under both clean and noisy misalignment conditions. In the clean setting, GraphBEV outperforms BEVFusion \cite{bevfusion-mit} significantly, while in the noisy setting, the performance improvement is substantial. Furthermore, it is evident that BEVFusion \cite{bevfusion-mit} exhibits a significant decrease in metrics such as mAP and NDS when transitioning from clean to noisy conditions, whereas GraphBEV demonstrates a smaller decrease in performance metrics.

Notably, adding the LocalAlign or GlobalAlign module to BEVFusion \cite{bevfusion-mit} has minimal impact on latency compared to BEVFusion \cite{bevfusion-mit} alone, and the latency is lower than that of TransFusion\cite{Transfusion}. When only the LocalAlign module is added to BEVFusion \cite{bevfusion-mit} and the KD-Tree algorithm is used to build proximity relationships, while fusing projected depth with neighbor depth to prevent feature misalignment, significant enhancements are observed in both clean and noisy misalignment settings. Adding only the GlobalAlign module to BEVFusion \cite{bevfusion-mit} also leads to noticeable improvements. Particularly, the simultaneous addition of the LocalAlign and GlobalAlign modules exhibits strong performance in both clean and noisy settings.

\begin{table*}[htp]
\scriptsize
\centering
  \caption{Effect of the Hyperparameters $K_{\text{graph}}$ for Feature Misalignment. We analyze the effect of hyperparameter $K_{\text{graph}}$ in LocalAlign module for feature alignment under \textcolor{red}{noisy} misalignment settings on the nuScenes  validatio set. `LT(ms)' represents latency. All latency measurements are conducted on the same workstation with an A100 GPU.}
  \renewcommand\arraystretch{1}
  \resizebox{\linewidth}{!}{
\begin{tabular}{ccc|ccc|ccc|ccc|ccc|ccc}
    \toprule
\multicolumn{3}{c|}{Baseline \cite{bevfusion-mit}}        
&\multicolumn{3}{c|}{$K_{\text{graph}}=5$}    
&\multicolumn{3}{c|}{$K_{\text{graph}}=8$} 
&\multicolumn{3}{c|}{$K_{\text{graph}}=12$} 
&\multicolumn{3}{c|}{$K_{\text{graph}}=16$} 
&\multicolumn{3}{c}{$K_{\text{graph}}=25$}  \\ 

mAP&NDS&LT 
&mAP&NDS&LT
&mAP&NDS&LT 
&mAP&NDS&LT 
&mAP&NDS&LT 
&mAP&NDS&LT \\
\midrule

60.8& 65.7& 132.9&     67.1& 70.9& 138.2&    69.1& 72.0& 141.0&    69.8& 72.2& 143.4&               
 68.8& 70.5& 145.3&    67.1& 69.9& 150.0
\\ 

\bottomrule
\end{tabular} }

\label{tab_nuscenes_K}
\end{table*}

\hspace{-1.3em}
\begin{minipage}[c]{0.37\textwidth}
\tiny
\centering
\captionof{table}{Robustness to weather conditions on nuScenes \cite{nuscenes} clean validation set. Notably, the evaluation metric is mAP. }
    \label{tab_nuscenes_robustness_weather}
  \renewcommand\arraystretch{1}
\begin{tabular}{l|cccc }
    \toprule
  \multirow{2}{*}{Method} &\multicolumn{4}{c}{  Different Weather Conditions }    \\
          & Sunny  & Rainy  & Day  & Night   \\ 
\midrule

Baseline \cite{bevfusion-mit}& 68.2& 69.9& 68.5& 42.8 \\ 
\textbf{GraphBEV}  &70.1&70.2& 69.7& 45.1\\ 

\bottomrule
\end{tabular} 
\end{minipage}
\hspace{0.3em}
\begin{minipage}[c]{0.58\textwidth}
\tiny
\centering
    \captionof{table}{Robustness to  different ego distances, different sizes on nuScenes \cite{nuscenes} clean validation set. Notably, the evaluation metric is mAP. }
\label{tab_nuscenes_ego_distance}      
  \renewcommand\arraystretch{1}
\begin{tabular}{l|ccc| ccc }
    \toprule
  \multirow{2}{*}{Method} &\multicolumn{3}{c|}{Different Ego Distances}&\multicolumn{3}{c}{Different Object Sizes}  \\
          & Near & Middle  & Far & Small & Moderate & Large  \\ 
\midrule
TransFusion-L \cite{Transfusion} & 77.5& 60.9& 34.8 & 44.7& 54.5& 60.4 \\ 
Baseline\cite{bevfusion-mit} & 79.4 &64.9& 40.0 & 50.3& 58.7& 64.0 \\ 
\textbf{GraphBEV} &78.6& 65.3& 42.1&55.4& 58.3& 63.1\\ 

\bottomrule
\end{tabular}
\end{minipage}

\subsubsection{Effect of the Hyperparameters $K_{\text{graph}}$ for Feature Misalignment.}

As shown in Table \ref{tab_nuscenes_K}, to analyze the impact of the hyperparameter $K_{\text{graph}}$ in the LocalAlign module on feature misalignment, we studied its effects under noisy misalignment settings on the nuScenes validation set. $K_{\text{graph}}$, which is the number of nearest depths for LiDAR-to-camera projected depth in the LocalAlign module, influences the expressive capability of neighboring depth features. It is observed that our GraphBEV achieves optimal overall performance when $K_{\text{graph}}$ is set to 8. Therefore, selecting an appropriate value for $K_{\text{graph}}$ is crucial and may vary across different datasets. Furthermore, despite significant fluctuations in mAP resulting from changes in $K_{\text{graph}}$, the overall performance still surpasses that of BEVFusion.

\subsubsection{Robustness to Weather Conditions.}

As shown in Table \ref{tab_nuscenes_robustness_weather}, we present a robustness analysis of our GraphBEV with respect to different weather conditions. Various weather conditions influence the 3D object detection task. Following the approach of BEVFusion\cite{bevfusion-mit} and ObjectFusion \cite{ObjectFusion}, we partition the scenes in the validation set into sunny, rainy, day, and night conditions. We outperform BEVFusion\cite{bevfusion-mit} under different weather conditions, especially in night scenes. Overall, our GraphBEV improves performance in sunny weather through accurate feature alignment and enhances performance in adverse weather conditions.
\subsubsection{Robustness to Ego Distances and Object Sizes.}
As shown in Table \ref{tab_nuscenes_ego_distance}, we analyzed the impact of different ego distances and object sizes on the performance of GraphBEV. We categorized annotation and prediction ego distances into three groups: Near (0-20m), Middle (20-30m), and Far (>30m), and summarized the size distributions for each category, defining three equal-proportion size levels: Small, Moderate, and Large. It is evident that GraphBEV demonstrates significant performance improvements for distant and small objects. Compared to BEVFusion\cite{bevfusion-mit}, our GraphBEV consistently enhances performance across all ego distances and object sizes, further narrowing the performance gaps. Overall, our GraphBEV exhibits greater robustness to changes in ego distances and object sizes.

\section{Conclusion}
In this work, we propose a robust fusion framework, GraphBEV, to address the feature misalignment issue for BEV-based methods. To mitigate feature misalignment caused by inaccurately projected depth provided by LiDAR, we propose the LocalAlign module, which utilizes neighbor-aware depth features through graph matching. Furthermore, to prevent global misalignment during the fusion of LiDAR and camera BEV features, we designed the GlobalAlign module to simulate offset noise, followed by aligning global multi-modal features through learnable offsets. Our GraphBEV significantly outperforms BEVFusion on the nuScenes validation set, especially in noisy misalignment settings. It demonstrates that our GraphBEV greatly advances the application of BEVFusion in real-world scenarios, particularly under noisy misalignment conditions.
\section*{Acknowledgements}
We sincerely appreciate the helpful discussions provided by Hongyu Pan from Horizon Robotics. This work was supported by the National Key R\&D Program of China (2018AAA0100302).
\bibliographystyle{splncs04}
\bibliography{main}
\end{document}